\documentclass[conference]{IEEEtran}
\usepackage{amsmath,epsfig}
\usepackage{cite}
\usepackage[T1]{fontenc}
\usepackage[utf8]{inputenc}
\usepackage{fancyhdr}
\usepackage{ifthen}
\usepackage{caption}
\usepackage{float}
\usepackage{amsmath}
\usepackage{ wasysym }
\usepackage{ amssymb }
\usepackage{ upgreek }%\usepackage[ruled,noend]{algorithm2e}
\usepackage{algorithm}
\usepackage{algpseudocode}
\usepackage{pifont}
\usepackage{tikz}
\usetikzlibrary{arrows,matrix,positioning}
\usepackage{amsthm}
\usepackage{enumerate}
\usepackage{subfigure}
\usepackage{graphicx}
\usepackage{bbm}
\usepackage{wrapfig}
\usepackage{cancel}
\usepackage[bookmarks=false]{hyperref}
\usepackage{pdfpages}
\usepackage{tabularx}
\usepackage{hyperref}
\usepackage{pgfplots}
\usepackage{color}
\usepackage{mathrsfs}
\usepackage{multirow}

\newtheorem*{theorem*}{Theorem}

\newcommand{\bpg}{\begin{paragraph}{}}
\newcommand{\epg}{\end{paragraph}}
\newcommand{\setpg}[1]{\begin{paragraph}{\bf Question #1:}}
\newcommand{\unsetpg}{\end{paragraph}\newpage}
\newcommand{\bit}{\begin{itemize}}
\newcommand{\eit}{\end{itemize}}
\newcommand{\beq}{\begin{equation}}
\newcommand{\eeq}{\end{equation}}
\newcommand{\beqn}{\begin{equation*}}
\newcommand{\eeqn}{\end{equation*}}
\newcommand{\beqa}{\begin{equation}\begin{aligned}}
\newcommand{\eeqa}{\end{aligned}\end{equation}}
\newcommand{\beqna}{\begin{equation*}\begin{aligned}}
\newcommand{\eeqna}{\end{aligned}\end{equation*}}

\newcommand{\ii}{\item}

\newcommand{\enum}{\begin{enumerate}}
\newcommand{\enuma}{\begin{enumerate}[(a)]}
\newcommand{\eenum}{\end{enumerate}}

\newcommand{\vphi}{\varphi}

\newcommand{\matrixcolsep}[1]{\kern#1em\vline}

\newcommand{\boly}{\boldsymbol{y}}

\definecolor{dkgreen}{rgb}{0,0.6,0}
\definecolor{gray}{rgb}{0.5,0.5,0.5}
\definecolor{mauve}{rgb}{0.58,0,0.82}

\algnewcommand\algorithmicinput{\textbf{Input}:}
\algnewcommand\INPUT{\item[\algorithmicinput]}
\algnewcommand\algorithmicoutput{\textbf{Output}:}
\algnewcommand\OUTPUT{\item[\algorithmicoutput]}

\title{What's Mine is Yours:  Pretrained CNNs for Limited Training Sonar ATR}

\author{\IEEEauthorblockN{John McKay$^\ast$, Isaac Gerg$^\square$, Vishal Monga$^\ast$, \& Raghu G. Raj$^\odot$}
\IEEEauthorblockA{$^\ast$Department of Electrical Engineering, 
Pennsylvania State University, 
State College, PA\\
$^\square$Applied Research Laboratory, 
Pennsylvania State University,
State College, PA\\
$^\odot$U. S. Naval Research Laboratory, Washington, DC}\thanks{J. McKay, V. Monga, \& R. Raj were supported by ONR Grant 0401531.}}
\IEEEoverridecommandlockouts 
\begin{document}
\maketitle

%% Abstract 
\begin{abstract}
Finding mines in Sonar imagery is a significant problem with a great deal of relevance for seafaring military and commercial endeavors.  Unfortunately, the lack of enormous Sonar image data sets has prevented automatic target recognition (ATR) algorithms from some of the same advances seen in other computer vision fields.  Namely, the boom in convolutional neural nets (CNNs) which have been able to achieve incredible results - even surpassing human actors - has not been an easily feasible route for many practitioners of Sonar ATR.  We demonstrate the power of one avenue to incorporating CNNs into Sonar ATR:  transfer learning.  We first show how well a straightforward, flexible CNN feature-extraction strategy can be used to obtain impressive if not state-of-the-art results.  Secondly, we propose a way to utilize the powerful transfer learning approach towards multiple instance target detection and identification within a provided synthetic aperture Sonar data set.
\end{abstract}

%% Peer Review maketitle
\IEEEpeerreviewmaketitle

%% *** Introduction
\section{Introduction}\label{sec:introduction}
Convolutional neural networks (CNNs) are the current ``go-to'' for just about every image classification problem.  Due in part to increasingly powerful and affordable GPUs, CNNs have been more and more utilized in research and to great effect; facial recognition \cite{schroff2015facenet}, hand writing analysis \cite{pham2014dropout}, and super resolution \cite{dong2014learning} are among a few of the topics that have seen incredible improvement because of CNNs.  Even the massively complex board game Go was not safe from the ever expanding dominion of CNNs and Deep Learning's success \cite{silver2016mastering}.  For many, they are becoming the starting point to any image classification problem - if they have the data.

Of course, that is the problem with CNNs:  they require quite a bit of data to fully train.  Even with dropout and up-sampling, having but a few hundred images may not suffice for a CNN with several hidden layers \cite{gal2015dropout}.  Despair not for there is a way around this:  transfer learning.  The creative minds in \cite{oquab2014learning} were among the first to realize that the information contained in the layers of a CNN may be useful for other problems.  Namely, the layers in models like VGGnet \cite{simonyan2014very} may - with tiny modification - be perfectly suitable for other image classification problems.  We do not control what exactly those layers yield feature-wise since they are not hand crafted but, still, they outperform SIFT and HOG features \cite{sharif2014cnn}.

For Sonar automatic target recognition (ATR), transfer learning is ideal.  There are no enormous, publicly available data sets that one can draw from online; Sonar ATR designers are stuck playing with limited training where even if they could get their hands on a cornucopia of images, very few could replicate their work.  That said, we show in the following paper that CNN features can be used in different ways to achieve Sonar image classification.  We show how useful CNN features can be to scan images and identify targets.

Specifically, we look to
\enum[i)]
\ii \textbf{Provide ample evidence that deep CNNs can be used as powerful feature extractors} and how, in combination with traditional support vector machines (SVMs), state-of-the-art results can be achieved.\label{item:svm} 
\ii \textbf{Outline a method for locating and identifying different types of mines} within a synthetic aperture Sonar (SAS) data set via a highly parallelizable and direct strategy.\label{item:new}
\ii \textbf{Reflect the robustness of CNN features to noise} in a manner relevant to Sonar ATR.  
\eenum
Section \ref{sec:PreviousWork} provides background for Sonar ATR, Section \ref{sec:UsingPretrainedCNNs} details how our two strategies work, and Section \ref{sec:Experiments} shows the impressive results on an actual Sonar image dataset provided by the Naval Surface Warfare Center.

%% ** Previous Work
\section{Previous Work}\label{sec:PreviousWork}
Research into Sonar ATR began in the 1980s and has grown roughly alongside general computer vision since then \cite{stack2011automation}.  There are a wide array of different strategies proposed for this problem.  Probably the most straight forward and well understood methods stem from SVMs applied to bulk feature extractions.  Algorithms like those presented in \cite{zhu2014model} focus on taking features stemming from a target's shape or shadow's behavior and then feed these into a standard SVM or similar classifier.  Another scheme that has seen some success with Sonar ATR is the sparse-reconstruction classification framework similar to \cite{wright2009robust}.  Methods such as \cite{mckay2015discriminative} take images and, without transformation, tries to classify images with an emphasis on resisting noise or occlusion.  Of course, there are in-betweens:  \cite{fandos2011optimal, kumar2012object} do feature transformations but then use SRC and \cite{mckay2016localized,mckay2016robust} try to use SRC but build a patch-based model to overcome geometric variability.

\begin{figure}[t]\centering
\includegraphics[width=1\columnwidth]{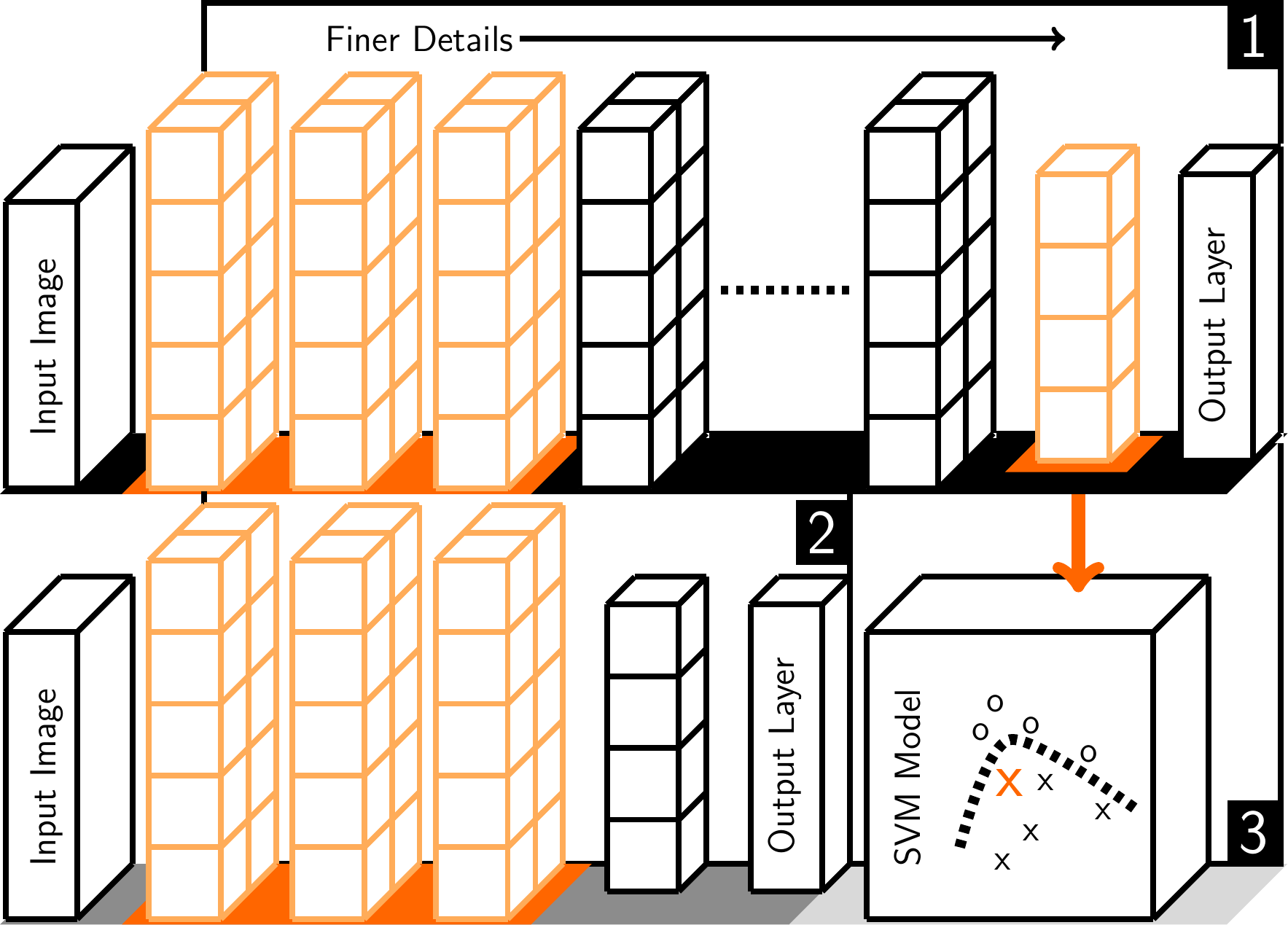}
\caption{Diagram of our two limited CNN strategies.  1) represents an original CNN with several heavily trained layers.  2) demonstrates the idea behind taking the first few layers incorporating it into a much smaller CNN.  3) displays the idea of using one of the latter layers as a feature transformation which is then fed into a SVM whose training also came from these transformations.}\label{fig:cnns}
\end{figure}

On the other hand, classifiers built off of pretrained CNNs have become more and more popular over time.  What do we mean by pretrained CNNs?  Many of the famous CNNs that have achieved the incredible results in image classification were trained from scratch using sizable datasets and several GPUs \cite{simonyan2014very,krizhevsky2012imagenet}.  A pretrained CNN uses weights from existing models - many times from those known CNNs - and either uses them straight (as we will with the SVM strategy) or tries to use them as a starting point that can lead to quicker training with smaller datasets \cite{oquab2014learning}.  The ability to harness the discriminative power of a CNN without the expensive training is alluring if not much more feasible for many applications.  Even  in cases where CNNs can be trained from scratch many use preset weights as it speeds up training and helps avoid issues with vanishing gradients \cite{wang2016cnn}.
 
 \begin{figure*}
    \centering
    \begin{minipage}[b]{1\linewidth}\centering
   	 \includegraphics[width=.24\columnwidth,trim={3cm 7cm 2cm 3cm},clip]{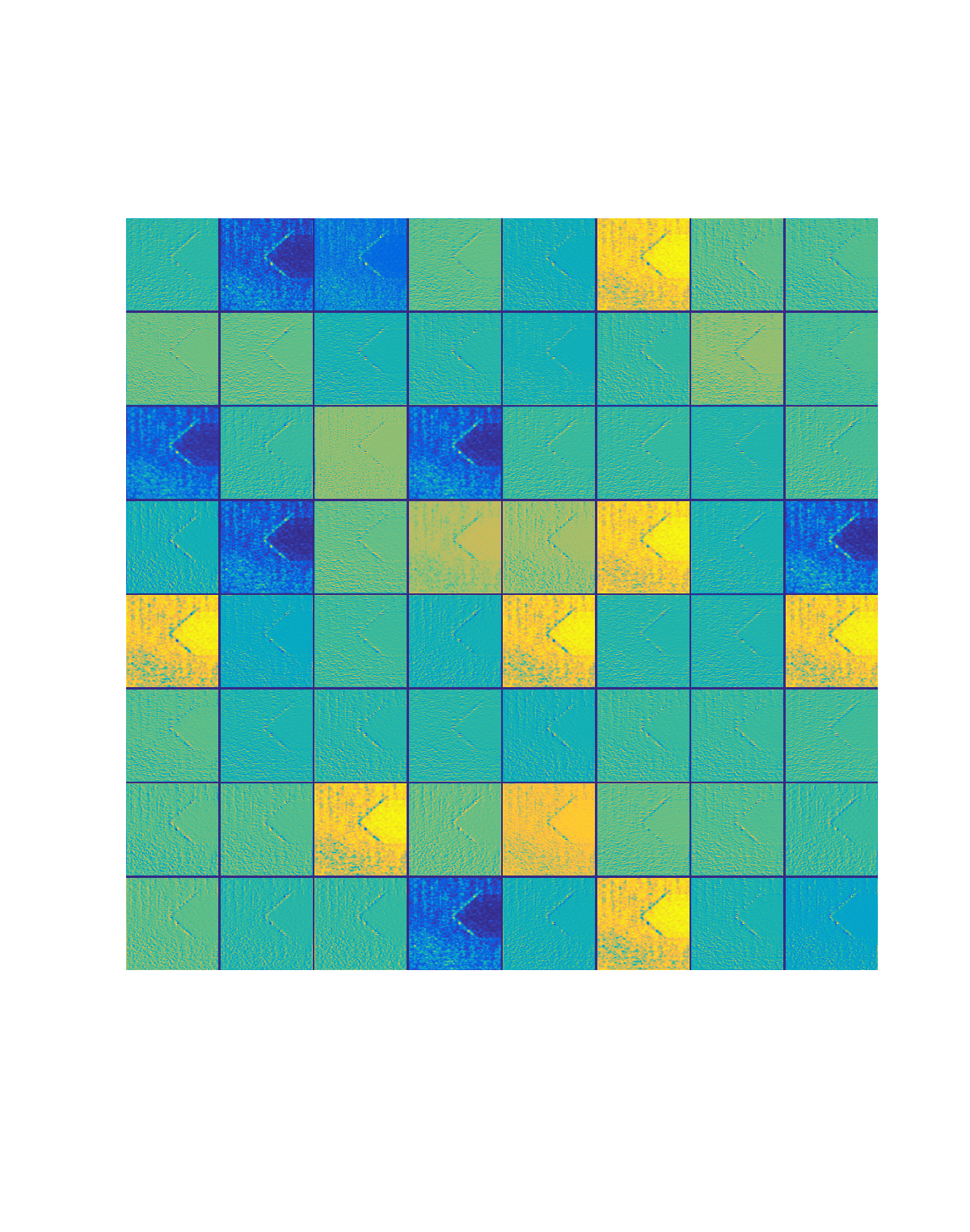}
  	  \includegraphics[width=.24\columnwidth,trim={3cm 7cm 2cm 3cm},clip]{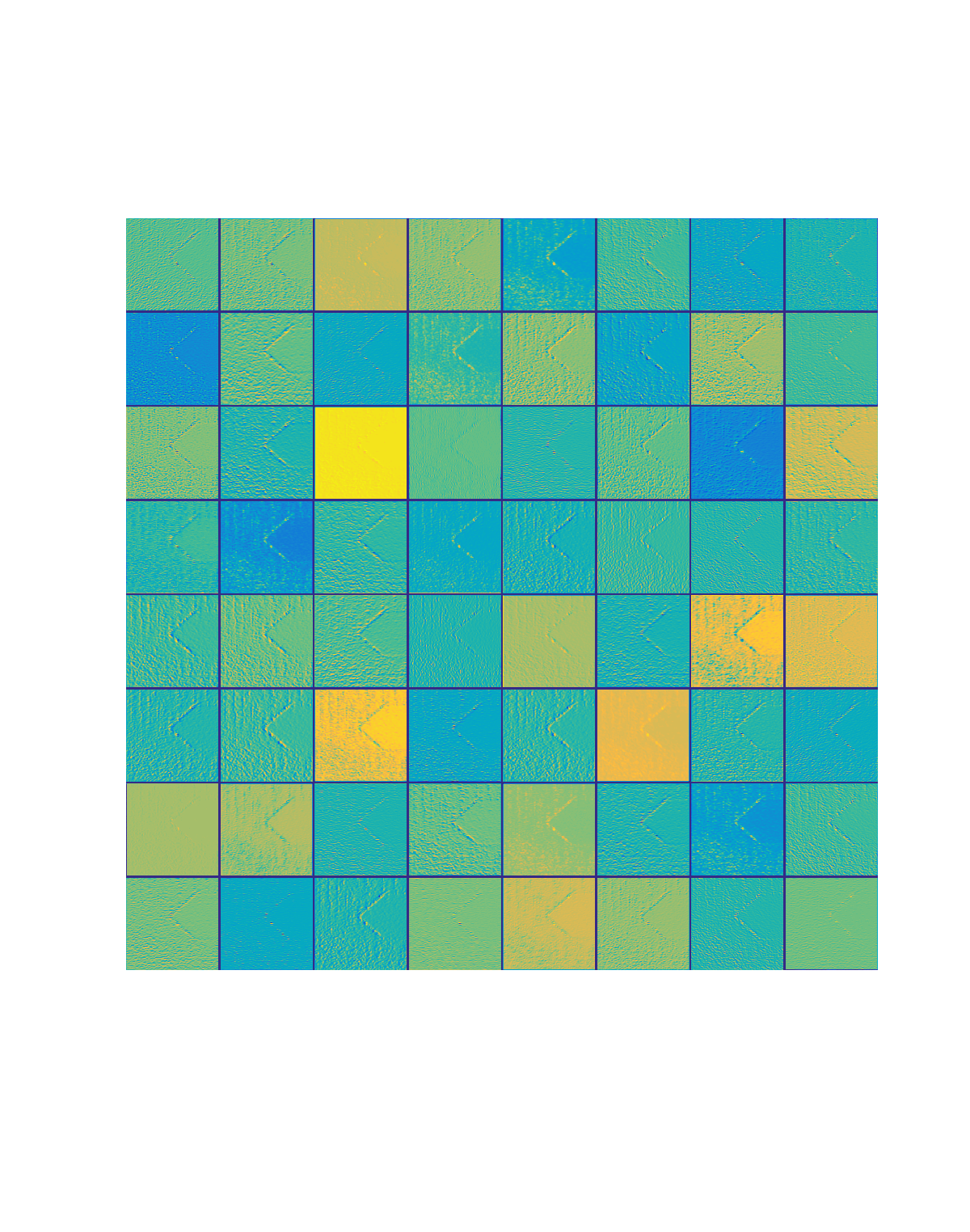}
 	   \includegraphics[width=.24\columnwidth,trim={3cm 7cm 2cm 6.5cm},clip]{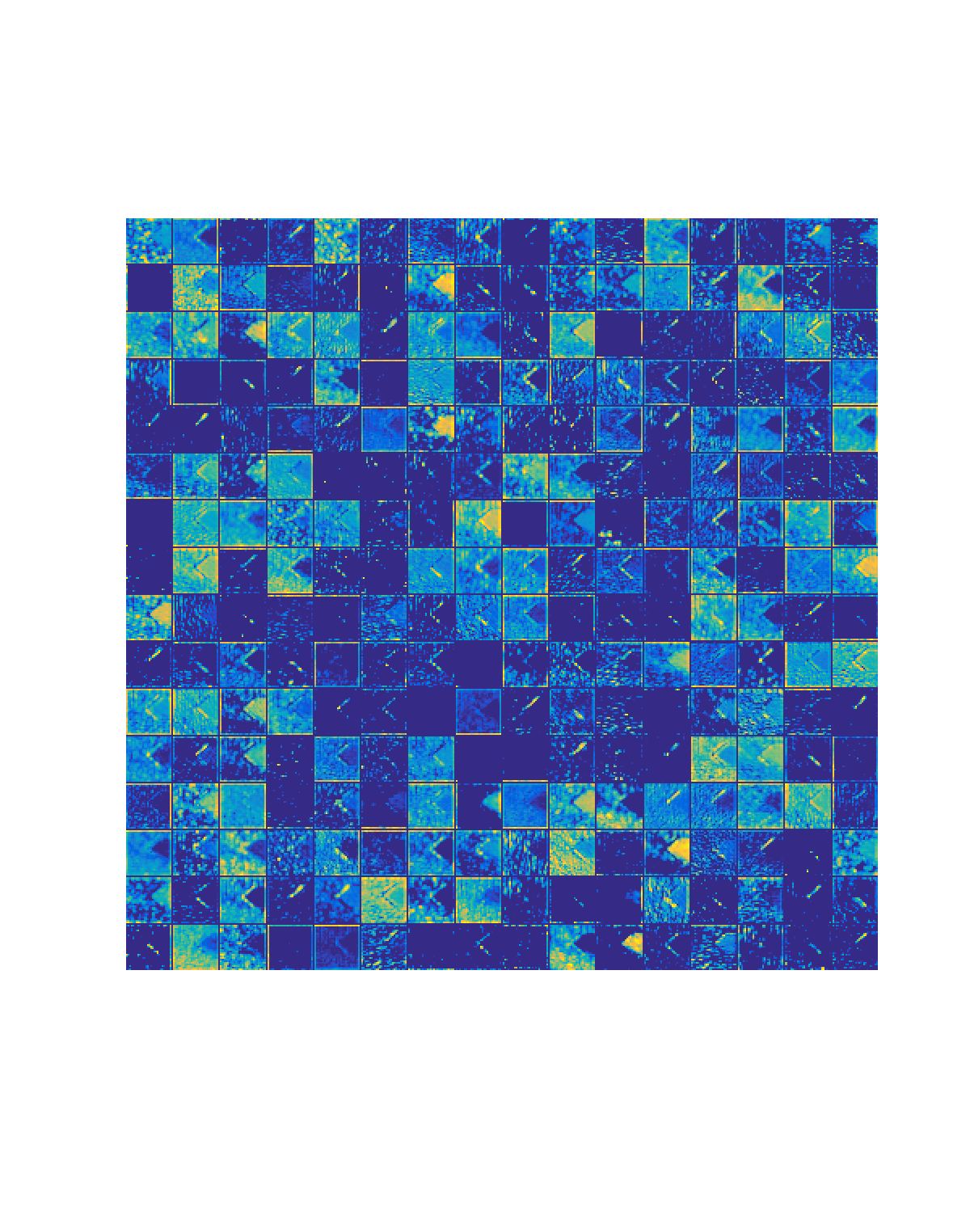}
	    \includegraphics[width=.24\columnwidth,trim={3cm 7cm 2cm 6.5cm},clip]{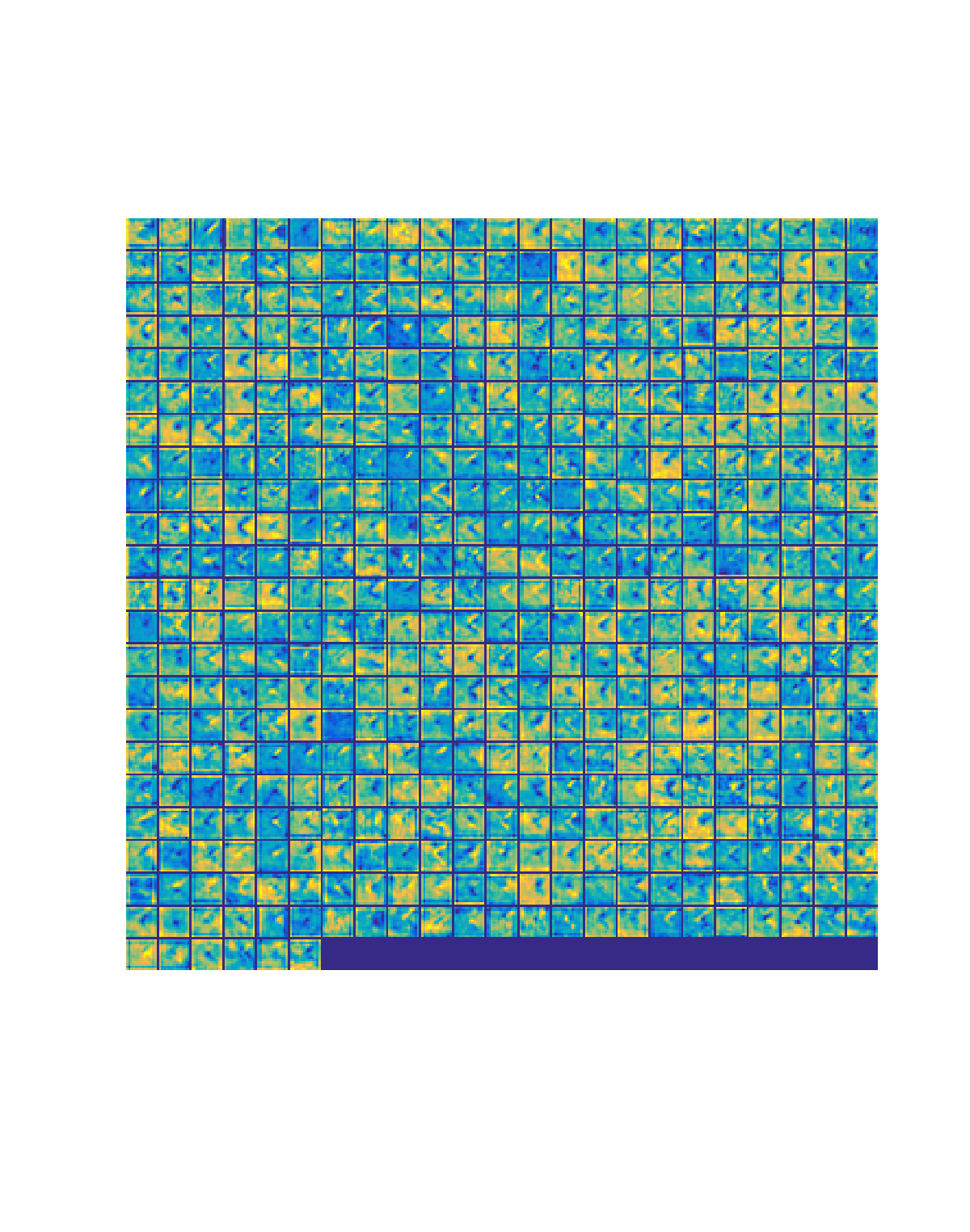}
	    \caption{Example Sonar image filter results through VGG19.  The top left comes from the first convolutional layer, top right from the second, bottom left from the fourth, and bottom right from the fifth.}
	    \label{fig:filterexamples}
    \end{minipage}    
    \begin{minipage}[b]{1\linewidth}\centering 
              \includegraphics[width=.49\columnwidth]{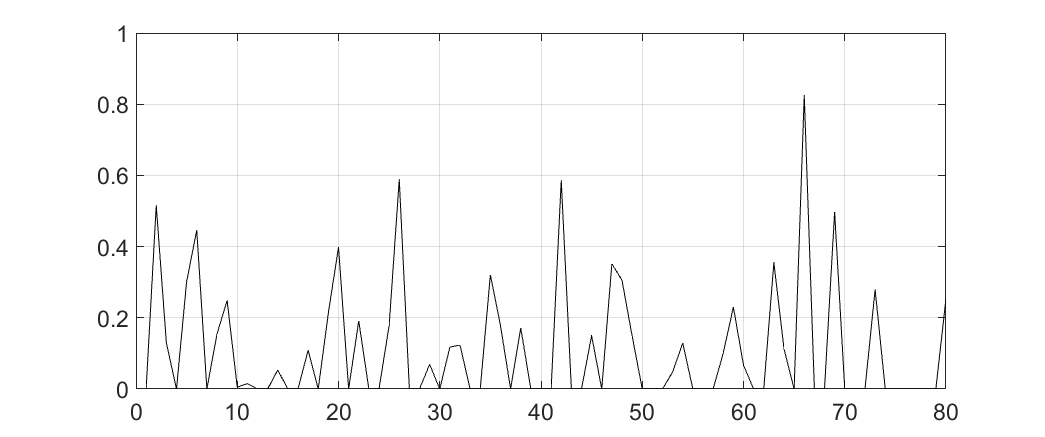}	
    	    \includegraphics[width=.49\columnwidth]{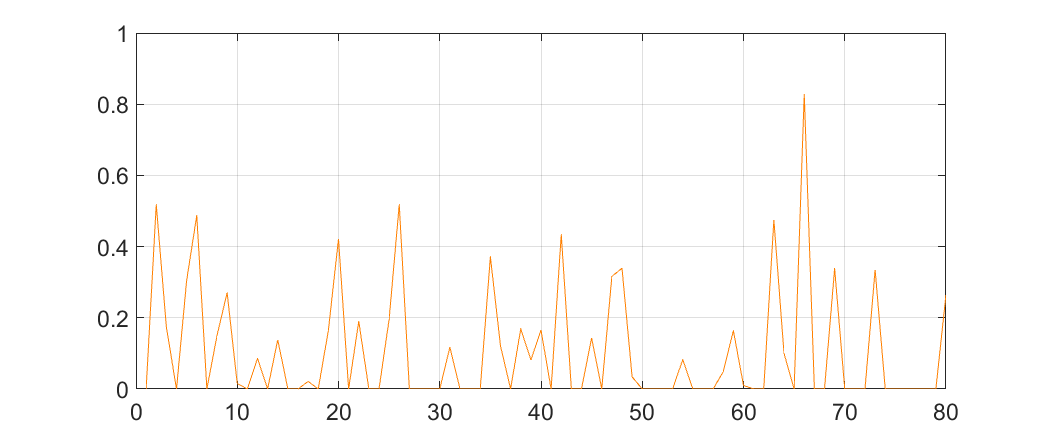}
    	    \includegraphics[width=.49\columnwidth]{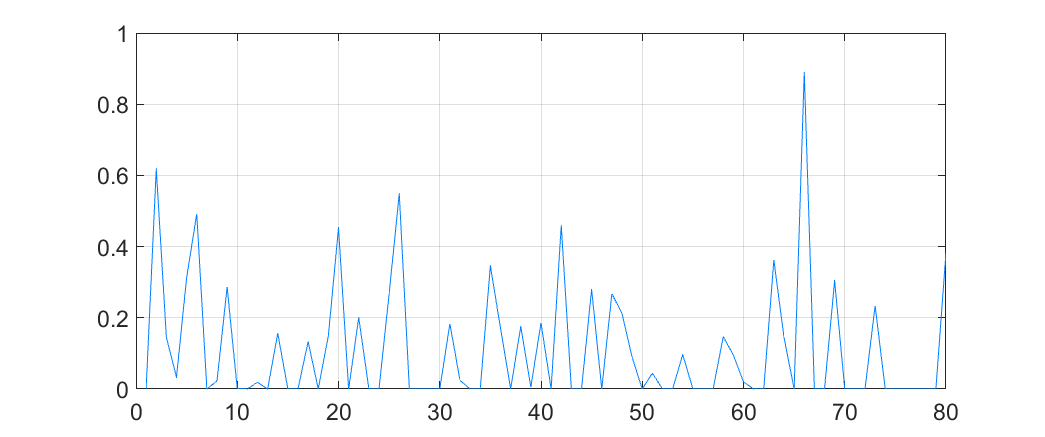}
    	    \includegraphics[width=.49\columnwidth]{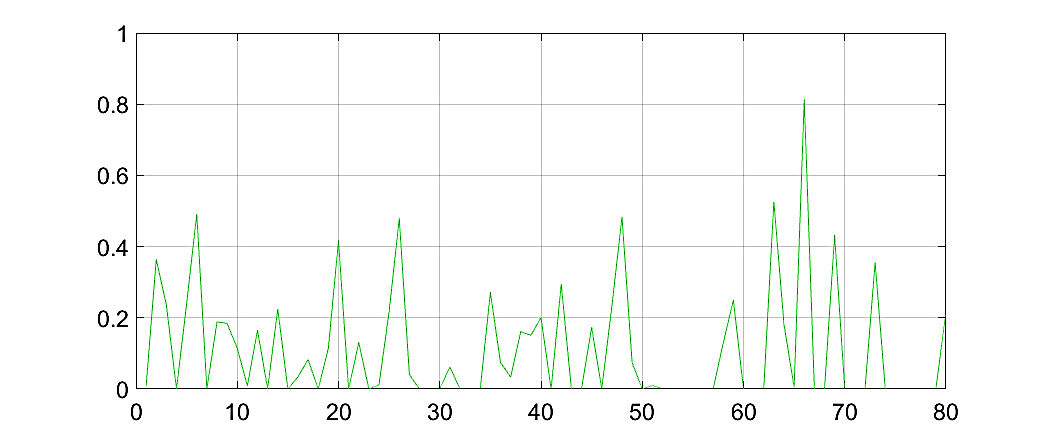}
    	    \caption{Example coefficient values for first 80 (of 4096) CNN Features from VGG19 of the four classes from RAWSAS dataset.  The classes represented are blocks (top left), cones (top right), spheres (bottom left), and cylinders (bottom right).}
    	    \label{fig:coefs}
    \end{minipage}
\end{figure*}
%% ** Using Pretrained CNNs
\section{Using Pretrained CNNs}\label{sec:UsingPretrainedCNNs}

Suppose you are given a sizable CNN with fixed weights, such as VGGnet.  We use $\boly_1^I,\dots,\boly_n^I$ to denote the layer outputs for the image $I$ that is put though the CNN.  How would you go about using it, let alone obtaining the fantastic results that others claim?  One way is to look at the final fully connected layers, $\boly_{n-k}^I,\dots,\boly_{n}^I$ and to treat one of these as transformed features.  For many multi-class models like VGGnet, the last layers are fully connected is the output of a softmax, the often referred to function $\vphi(x_i)=\exp(x_i)/\sum_j\exp(x_j)$.  As \cite{sharif2014cnn} suggests, this may not be suitable so $y_{n-1}$ - a vector output since it has to go into the softmax - serves as a worthy candidate.  Thus, we can proceed with a strategy similar to a typical SVM problem: using labeled $\boly_{n-1}^{I_1},\dots,\boly_{n-1}^{I_K}$ corresponding to the CNN feature extraction of images $I_1,\dots,I_K$ to train some function (linear, kernel, etc.) $f$, finding the transformation $\boly^T_{n-1}$ for a test image $T$, and then use $f(T)$ to figure out the label.  This is simple and, as we see in later sections, powerful.

Now, to those who are familiar with the work of \cite{oquab2014learning} and other transfer learning strategies for CNNs, the CNN feature extraction-SVM model may not be the first one to come to mind.  Indeed, in \cite{oquab2014learning}, the authors suggest \emph{fine-tuning} the weights of an existing neural network.  The idea here is that a CNN trained to decipher cars, flowers, and whatnot has generally captured a discriminatory set of weights that, with minimal tweaking, can be adapted to an unknown plethora of applications.  Shapes are shapes, after all, and models that can achieve impressive results on hundreds of different classes should be capable in telling the difference between, say, Sonar targets.  Thus, given a tiny set of images, one can use another, existing model as a starting point and then use the typical back-propagation-stochastic gradient procedure to do the fine tuning.  The original CNN has the final layer designed according to the number of classes.  When properly fine-tuning a model to a new problem (like Sonar ATR), what one has to do is simply replace it with a new (softmax in our case) layer that gets trained as the model is fine tuned.

For deeper CNNs like \cite{simonyan2014very} or \cite{krizhevsky2012imagenet}, the process of understanding finer and finer details goes accordingly to the layers; while the first layers may mimic simple edge filters, the later ones tend to capture complexities like the grill of a car \cite{lee2011unsupervised}.  For Sonar, this means that later layers may \emph{not} be that useful.  Depending on the setting, certain targets may not require the deepest layers of a larger model.  Thus, when taking the fine-tuning route, many need only utilize a portion of the initial layers as to include the level of detail needed.

\section{Experiments}\label{sec:Experiments}

Now that we have built up our pretrained CNN methods, let's see how they do on actual Sonar data.  The Naval Surface Warfare Center provided us with a four-class data set of synthetic aperture Sonar images.  Our problem is to discern between the images of blocks, cones, spheres, and cylinders (several example images from this data set can be found in \cite{mckay2015discriminative,mckay2016localized,mckay2016robust}) with sixty examples per class.  The blocks and cylinders can be difficult for even a person to discern between, so the data is fairly nontrivial in that way, not to mention the considerable background clutter and that our samples included some positional jitter.  

\subsection{Target Recognition}

With this data set, we designed four trials per method using 20 training images per class and then tested on a random assortment of ten examples per class.  In addition to our two pretrained CNN strategies, we used a straight forward SIFT bag-of-words model (BOW) and the SRC method of \cite{mckay2015discriminative}.  To measure effectiveness, we used  precision, a metric that shows how confident one can be that a given classification is correct, and recall, a value of how well a model correctly classifies a target.  Formally, consider for any class the sets of tested images that are correctly attributed to that class (true positive-$TP$), identified as that class but actually belonging to another (false positive-$FP$), and belonging to that class but identified to another (false negative-$FN$); the precision metric for a class is then the ratio $TP/(TP+FP)$ and the recall is $TP/(TP+FN)$.

We tested several different existing, pretrained CNNs for our SVM and modified CNN problems:  VGGnet with depth 16 (VGG16), VGGnet with depth 19 (VGG19) \cite{simonyan2014very}, fast VGGnet (VGG-f) \cite{chatfield12014return}, and Alexnet (Alex) \cite{krizhevsky2012imagenet}.  We used MatConvNet \cite{vedaldi2015matconvnet} from which the reader can replicate our results.

\begin{figure}[t]\centering
\includegraphics[width=1\columnwidth]{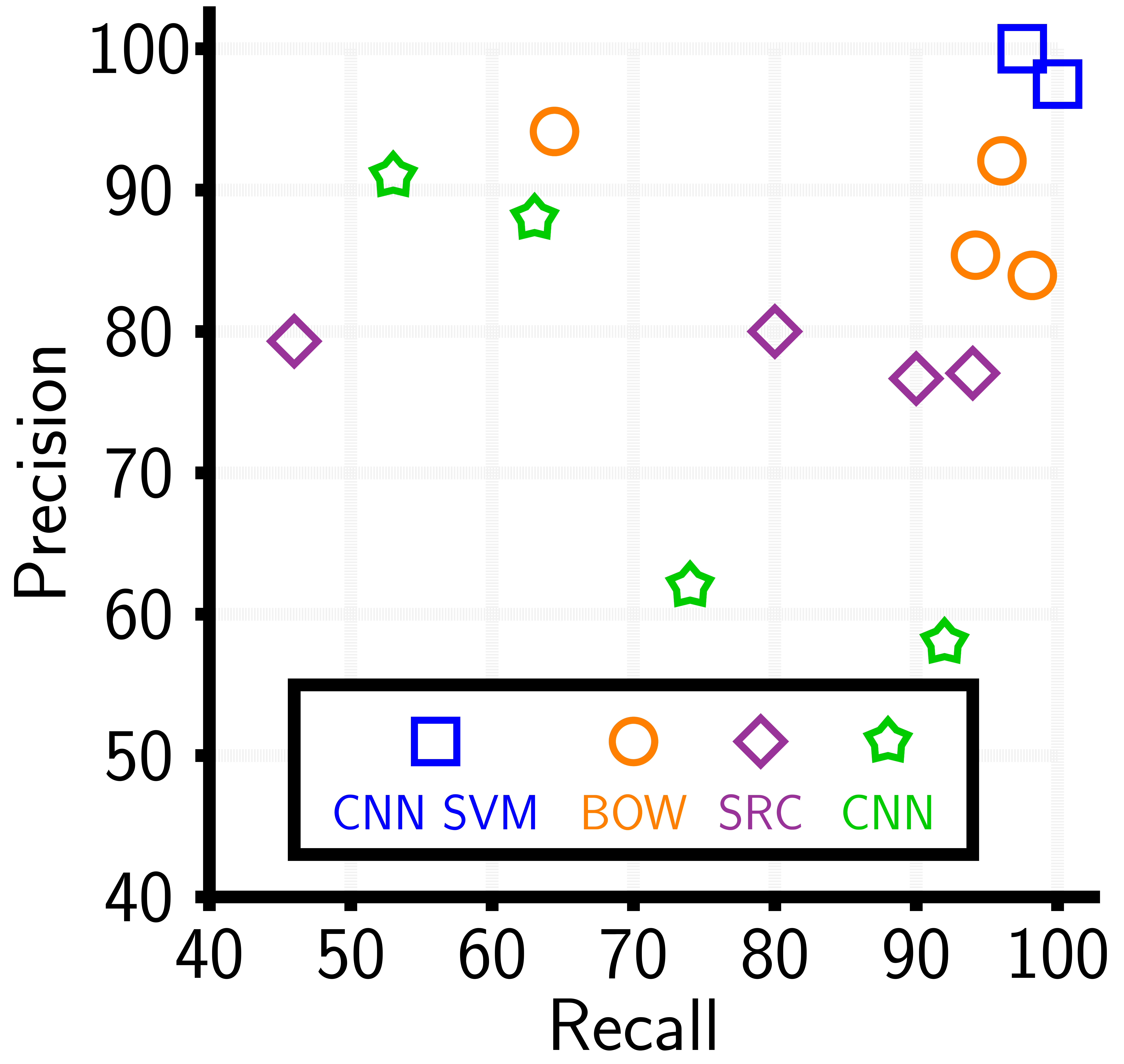}
	    \begin{tabular}{|c|c|c|c|c|}\hline
	    \textbf{Mean Values (\%)} & CNN SVM & BOW & SRC & CNN\\\hline
	    Precision & 98.8 & 88.9 & 78.3 & 74.8\\\hline
	    Recall & 98.8 & 88.2 & 77.5 & 70.5\\\hline
	    \end{tabular}
\caption{Precision-recall values for the four class problem according to the SVM model using Alexnet as a feature extractor (CNN SVM), SIFT-feature SVM bag-of-words (BOW), sparse reconstruction-based classifier (SRC), and the fine-tuned CNN constructed from the first three convolutional layers of VGG-f (CNN).  Note that blocks overlapped with cylinders and spheres overlapped cones for the CNN SVM.}\label{fig:pr}
\end{figure}

\begin{figure}[t]\centering
\includegraphics[width=1\columnwidth]{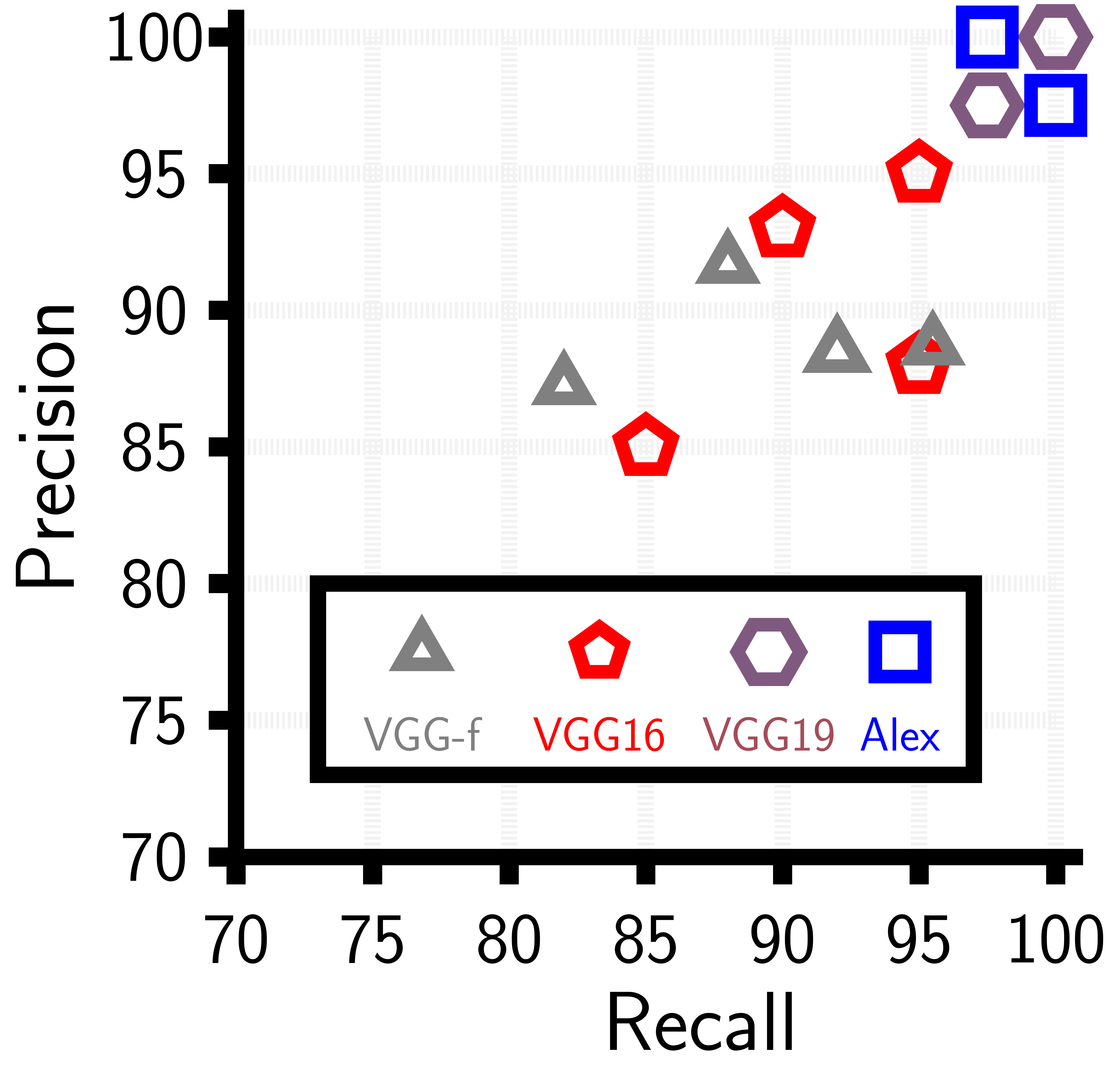}
	    \begin{tabular}{|c|c|c|c|c|}\hline
	    \textbf{Mean Values (\%)} & VGG-f & VGG16 & VGG19 & Alex\\\hline
	    Precision  & 89.1 & 90.3 & 98.8 & 98.8\\\hline
	    Recall & 89.4 & 91.3 & 98.8 & 98.8\\\hline
	    \end{tabular}
\caption{Per class precision and recall values for VGG-f, VGG16, VGG19, and Alexnet.  Note that, similar to Figure \ref{fig:pr}, VGG19 overlaps with itself as does Alexnet.  For Alex, blocks over spheres, cones over cylinders.}\label{fig:prsvm}
\end{figure}

First, to get an idea how how these CNNs handled the Sonar data, refer to Figure \ref{fig:filterexamples} where we break down the layer activations of VGG19 (the original model, not fine-tuned) with a block input.  There, the increasingly discriminative nature of the filters reveals an increasing attempt to understand complex details.  The first couple layers look somewhat like the output of traditional edge filters (especially the second layer).  The later ones, though, reveal an almost bizarre understanding of the input image with several different highlighted aspects with no obvious interpretation, though we can see the different types of illumination and parsing that the model is trying do.  Note that if we look at the activated layers after an optical image of say a dog or car is pass through the model, we would see emergent details in the later layers such as details pertaining to the nose of the dog or mesh-like shapes from the grill of the car \cite{zeiler2014visualizing}.  Now, it is clear that our Sonar images do not share the level of resolution of that seen in the type of training sets used for the original VGG19 CNN, but it worth noting that the model does not involve features associated with the type of shadowing common to Sonar or high intensity differences between targets and background.  Basically, the CNN itself does not have a detailed nature for specifically Sonar and, as we will see, this is not something that can be easily fine-tuned into the model - especially not with as few data points as we have.  The features it extracts are still powerful but there is reason to be skeptical of CNNs trained for optical settings.

Figure \ref{fig:pr} shows the precision and recall values for each method according to each of the four classes (with some overlap) along with the values averaged across all classes.  It is readily apparent that the SVM using CNN features outperformed the two base line methods and the fine-tuned CNN by a bit.  This suggests that, while the CNN itself is unable to fine tune itself with such limited data (to capture the shadows and whatnot that we indicated) in order to achieve compelling results, the features themselves are highly discriminative elements.  When they are combined with a SVM they can overcome the limited training towards impressive results (the SVM with CNN features from Alexnet and VGG19 both had only two incorrect classifications out of 160 tests, each).  This hypothesis is furthered by the fact that the deeper and therefore more detailed models performed \emph{better} when combined with a SVM.  Figure \ref{fig:prsvm} shows the results of different SVM models and, while they all did relatively well when compared to the baseline methods and fine tuned CNN, there is a clear trend:  the more layers, the better the performance.  

All in all, these experiments show that Sonar ATR using existing CNN models has promise - especially with deeper CNNs.  We are curious to see if a fine-tuning model can be trained to learn more Sonar-specific qualities if given more data or if an entirely new model is needed (i.e. training a CNN from scratch).  This could have further implications for Sonar ATR as CNNs become even more prevalent.

%% ** Noisy Sonar Image Detection
\subsection{Target Detection with Noise}\label{sec:noise}

Noise is an unavoidable issue in many real-world imaging settings and Sonar is no different.  Of the places where one can find issues with noise is target detection wherein images are parsed to find mines.  The previous experiment we detailed involved the results of a target detection scheme, meaning that a detection scheme is what one would use to get the target chips that are then deciphered for their contents.  Can CNN features be used for detection and, therefore, an all-encompassing ATR strategy?  We present the following work to suggest that a response of yes.  Note that Gaussian noise models are ill suited for Sonar given how these images are formed so we, here, tested our detection scheme on images corrupted with Rayleigh-type noise \cite{mckay2016robust}.   

The basic idea is simple:  use the same CNN feature extraction (using Alexnet) and SVM discriminator from the previous experiment on several, uniformly selected \emph{overlapping} patches extracted from larger scheme.  Of course, the model is trained to discern between different targets - but - SVMs provide scores that indicate whether a test subject belongs to all the trained-upon classes.  If CNN features are particularly discriminative, it stands to reason that this would be reflected in the SVM output values and that a background image would cause some confusion, meaning lesser relative scores.  Thus, a simple threshold on the maximum score serves as a qualifier; if the SVM output's maximal value for a class is below $\uptau>0$, then the patch is deemed not to have a target; otherwise, it gets flagged.  In practice, we decided our threshold by a trial-and-error approach on a validation set (separate from the test image) with a tendency to have it on the lower end as to avoid false negatives.

\begin{figure*}\centering
	\includegraphics[width=.325\linewidth]{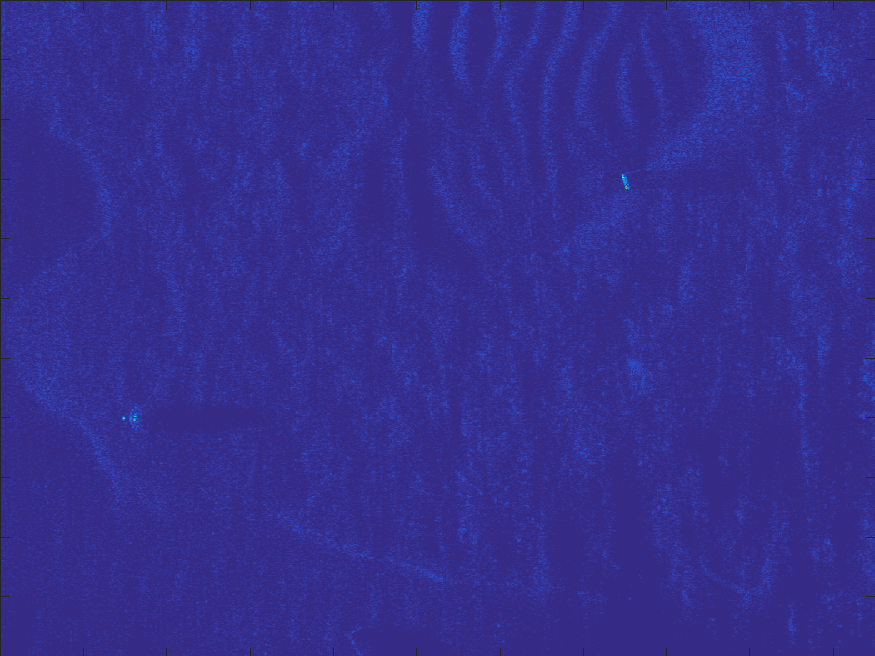}
	\includegraphics[width=.325\linewidth]{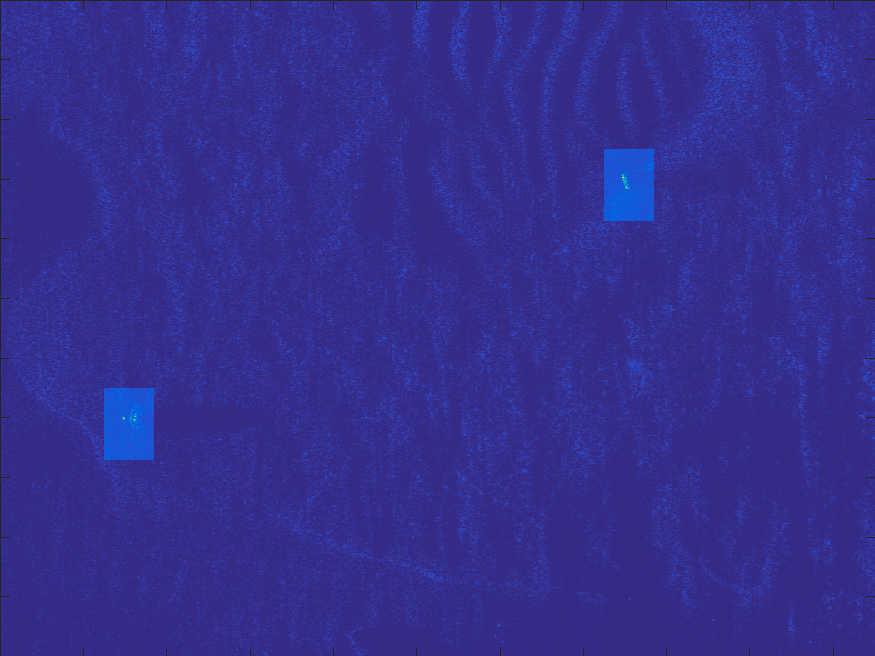}
	\includegraphics[width=.325\linewidth]{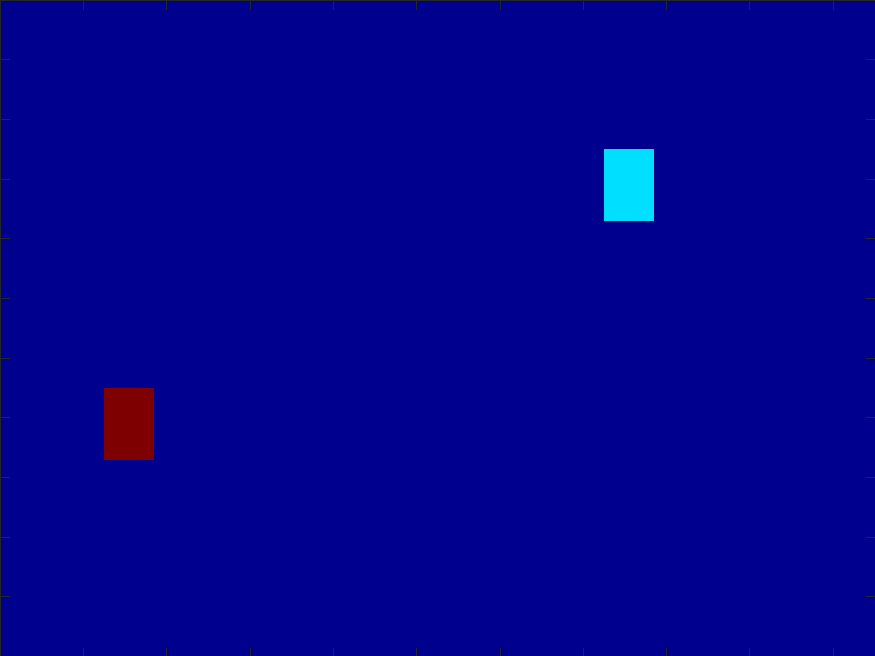}
    \caption{Target detection of two targets, a block and sphere, using our CNN feature extraction scheme.  The left-most image is the original magnitude scene, the middle has the areas highlighted where detection occurred, and the right has just the detection where the dark blue indicates areas where the associated patch didn't pass a threshold, red means that a patch passed a threshold and was classified a sphere, and the light blue shows patches that passed a threshold and were classified as a block.  Note that the threshold used (.9) was the same for all classes and found in a cross-validation stage prior to this formal detection.}\label{fig:twoscene}
\end{figure*}

We took to two scenarios to illustrate our detection scheme:  a case involving a scene with two targets and another involving one but with increasing Rayleigh noise intensity.  The first experiment with the two targets is show in Figure \ref{fig:twoscene} where one can see the background environment that could trip-up a detection.  Even still, using a threshold value of .9 devised during a cross-validation stage we used to tune our model, only two patches survived to indicate an object detection:  one overlaying the sphere-like mine and the other over the block-like mine.  In both cases the model also correctly identified the patches as being associated with their classes, meaning that no further classification method would have to be used in this case to classify detected targets.  

\begin{figure*}\centering
	    \begin{minipage}[b]{1\linewidth}\centering
	           \textbf{No Noise}\hspace{4cm}\textbf{PSNR: 35dB}\hspace{4.2cm}\textbf{PSNR: 25dB}\\\vspace{.05cm}
	    	\includegraphics[width=.32\columnwidth]{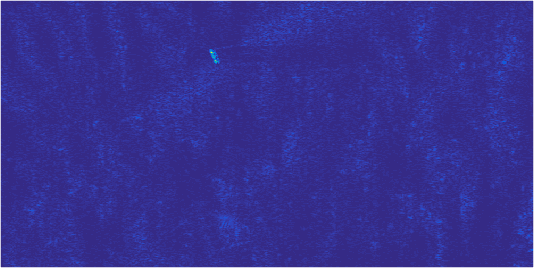}
	    	\includegraphics[width=.32\columnwidth]{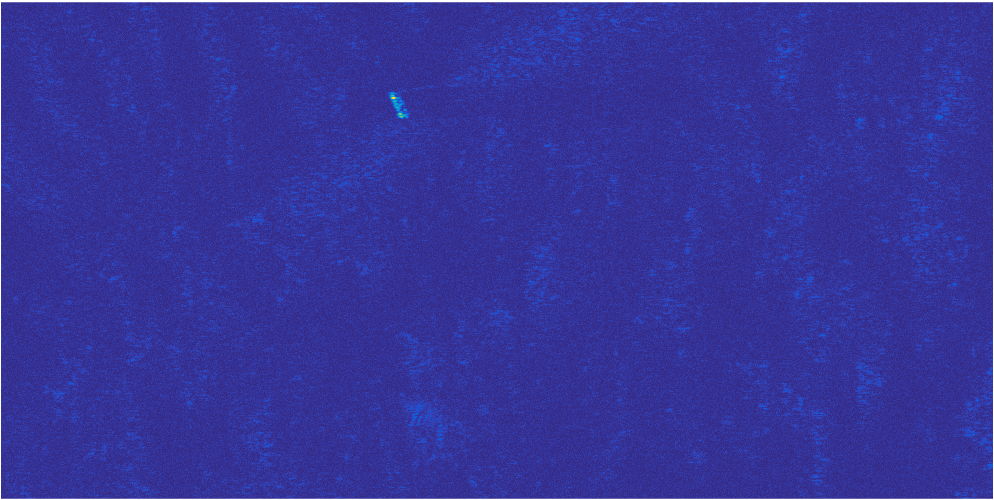}
	    	\includegraphics[width=.32\columnwidth]{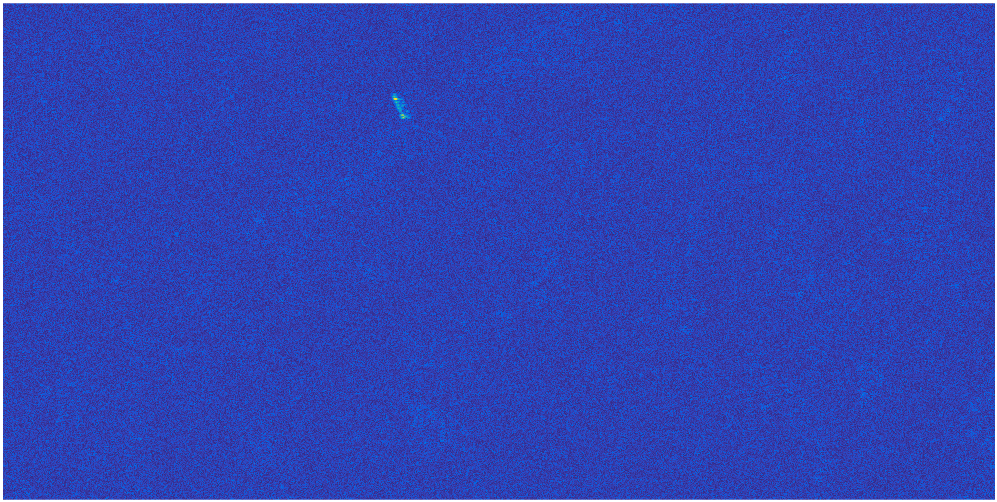}
	    \end{minipage}\vspace{.1cm}
	    \begin{minipage}[b]{1\linewidth}\centering
	    	\includegraphics[width=.32\columnwidth]{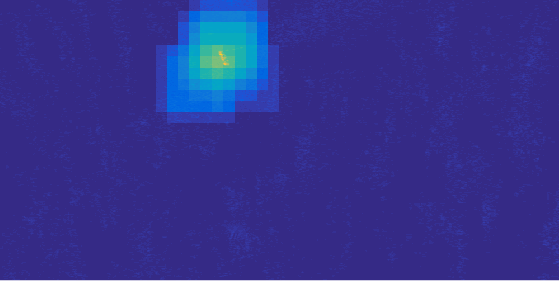}
	    	\includegraphics[width=.32\columnwidth]{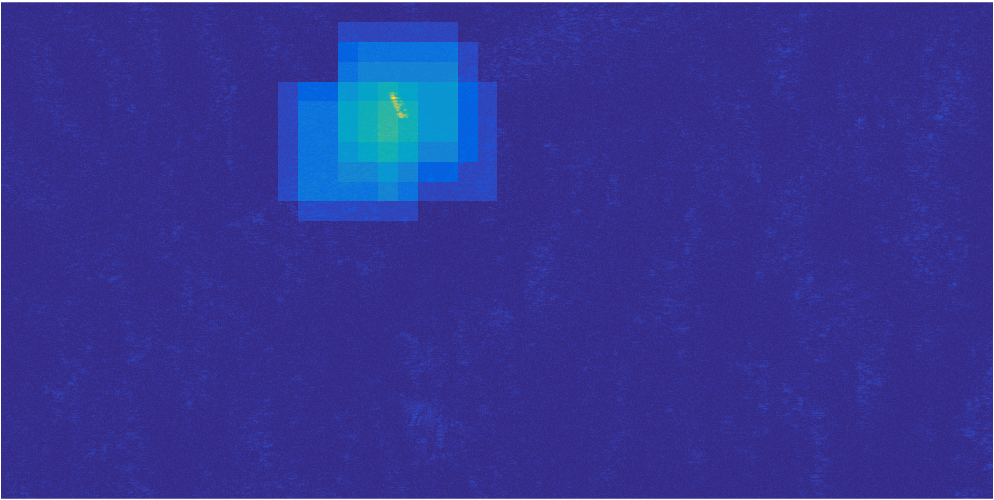}
	    	\includegraphics[width=.32\columnwidth]{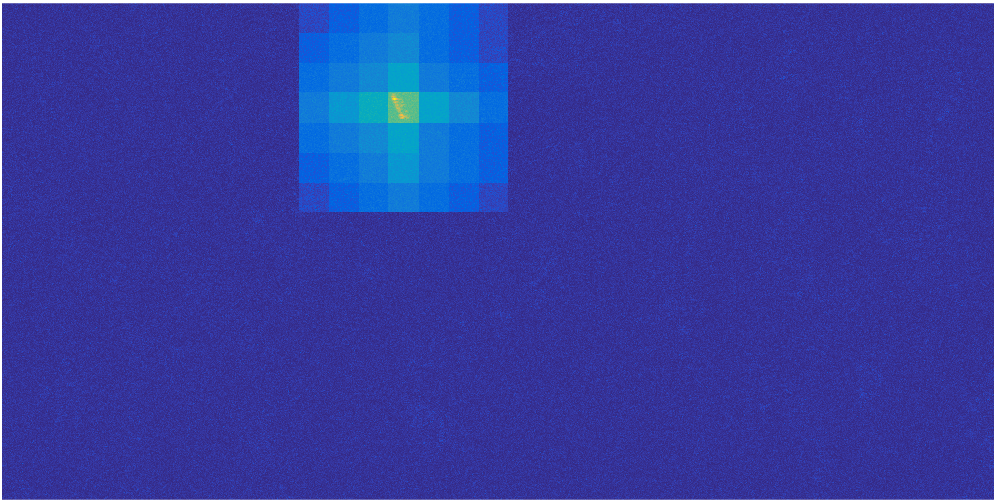}
	    \end{minipage}\vspace{.1cm}
	    \begin{minipage}[b]{1\linewidth}\centering
	    	\includegraphics[width=.32\columnwidth]{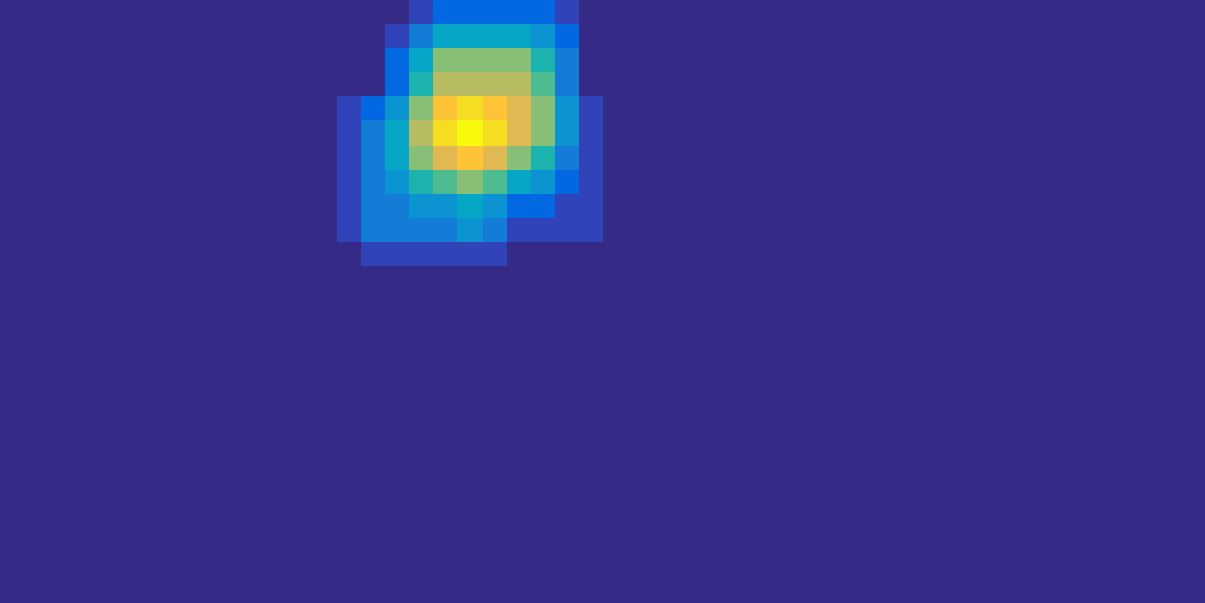}
	    	\includegraphics[width=.32\columnwidth]{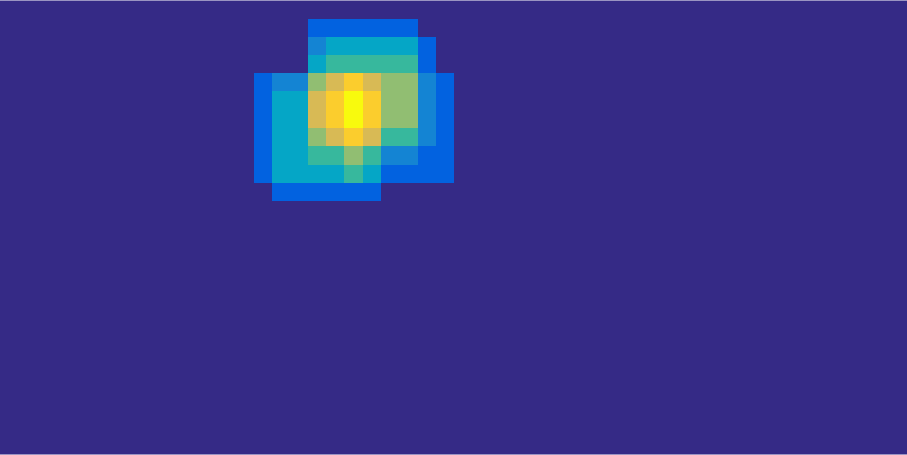}
	    	\includegraphics[width=.32\columnwidth]{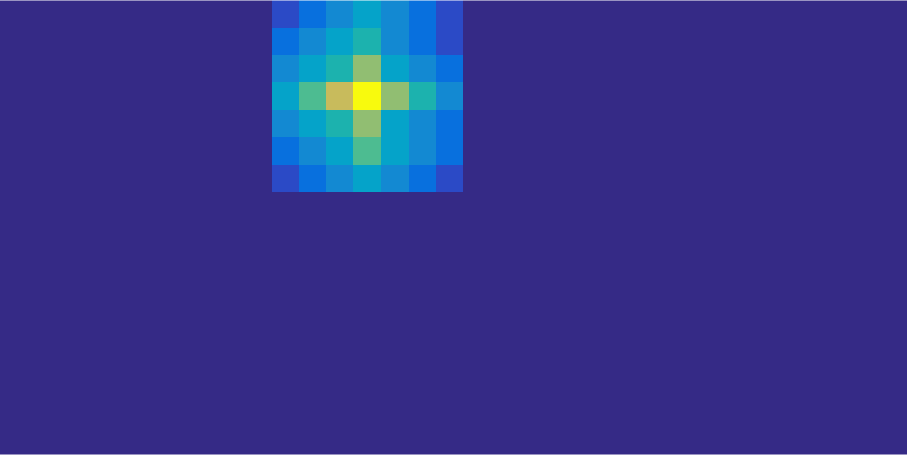}
	    \end{minipage}\vspace{.1cm}
	    \includegraphics[width=.7\columnwidth]{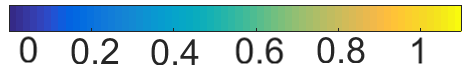}
    \caption{Target (block) detection results over three noise scenarios.  The top row shows the original image, the middle has the highlighted patches that passed a confidence threshold for target detection, and the bottom shows just the detected regions.  The threshold values for the normalized SVM output for detection was .9 in each case.}\label{fig:sceneWithNoise}
\end{figure*}

For the single-target case, Figure \ref{fig:sceneWithNoise} shows the results.  There, one can see the original image and the two corrupted versions along with the highlighted patches that passed a threshold indicating a target.  Overall, the method is able to overcome background clutter in the scene and identify the area wherein the target lies.  We suppose that a user could use the higher intensity regions (indicated best in the bottom row) to approximate where a target lies.  Note that in the case with the most noise, the edge of the block that is still able to come through despite the corruption serves as the main indicator for the target's detection.  In every patch highlighted to have passed the threshold, the classifier correctly deemed it to contain a block (which is unsurprising given the stellar results in the previous section).

In further work, it would be prudent to consider a SVM built not to decipher between classes but between targets and background.  While we have shown compelling results here, our method did struggle in larger scenes where more considerable background clutter can be confusing.  Given the powerful nature of CNN features and the fact that our method is amendable to limited training cases, a model could be designed with several different types of background clutter used for training and may be able to provide a relatively quick and discriminative detection scheme for large-scale Sonar targets.

%% *** Discussion
\section{Discussion}\label{sec:discussion}

CNNs are incredibly powerful tools that have found great success in many fields.  We presented a strategy that harnesses that discriminative abilities of existing CNNs towards small sample sonar ATR settings.  Our method is straightforward and powerful for both target recognition and detection.  We suggest that, like the authors of \cite{sharif2014cnn} for optical images, our strategy be used as a baseline; a relatively easy and fast algorithm that can be expected to do well on a diverse array of sonar datasets.  If a more sophisticated can out perform our own, then we know that it has more discriminatory ability than highly trained CNN features.  

As more and more models get published for classification, we see the potential for SVM models based on CNN features to grow.  Future work concerning our strategy should incorporate such new CNN features and try to tackle large-scale sonar scenes with several confusing background elements.  Our dataset lacked rocky patches or areas with heavy underwater vegetation; more work can be done to see how our approach can be tailored to such an environment.

%% *** Bibliography
\bibliographystyle{IEEEtran}
\bibliography{refIGARSS2017}

\begin{thebibliography}{10}
\providecommand{\url}[1]{#1}
\csname url@samestyle\endcsname
\providecommand{\newblock}{\relax}
\providecommand{\bibinfo}[2]{#2}
\providecommand{\BIBentrySTDinterwordspacing}{\spaceskip=0pt\relax}
\providecommand{\BIBentryALTinterwordstretchfactor}{4}
\providecommand{\BIBentryALTinterwordspacing}{\spaceskip=\fontdimen2\font plus
\BIBentryALTinterwordstretchfactor\fontdimen3\font minus
  \fontdimen4\font\relax}
\providecommand{\BIBforeignlanguage}[2]{{%
\expandafter\ifx\csname l@#1\endcsname\relax
\typeout{** WARNING: IEEEtran.bst: No hyphenation pattern has been}%
\typeout{** loaded for the language `#1'. Using the pattern for}%
\typeout{** the default language instead.}%
\else
\language=\csname l@#1\endcsname
\fi
#2}}
\providecommand{\BIBdecl}{\relax}
\BIBdecl

\bibitem{schroff2015facenet}
F.~Schroff, D.~Kalenichenko, and J.~Philbin, ``Facenet: A unified embedding for
  face recognition and clustering,'' in \emph{Proceedings of the IEEE
  Conference on Computer Vision and Pattern Recognition}, 2015, pp. 815--823.

\bibitem{pham2014dropout}
V.~Pham, T.~Bluche, C.~Kermorvant, and J.~Louradour, ``Dropout improves
  recurrent neural networks for handwriting recognition,'' in \emph{Frontiers
  in Handwriting Recognition (ICFHR), 2014 14th International Conference
  on}.\hskip 1em plus 0.5em minus 0.4em\relax IEEE, 2014, pp. 285--290.

\bibitem{dong2014learning}
C.~Dong, C.~C. Loy, K.~He, and X.~Tang, ``Learning a deep convolutional network
  for image super-resolution,'' in \emph{European Conference on Computer
  Vision}.\hskip 1em plus 0.5em minus 0.4em\relax Springer, 2014, pp. 184--199.

\bibitem{silver2016mastering}
D.~Silver, A.~Huang, C.~J. Maddison, A.~Guez, L.~Sifre, G.~Van Den~Driessche,
  J.~Schrittwieser, I.~Antonoglou, V.~Panneershelvam, M.~Lanctot \emph{et~al.},
  ``Mastering the game of go with deep neural networks and tree search,''
  \emph{Nature}, vol. 529, no. 7587, pp. 484--489, 2016.

\bibitem{gal2015dropout}
Y.~Gal and Z.~Ghahramani, ``Dropout as a bayesian approximation: Representing
  model uncertainty in deep learning,'' \emph{arXiv preprint arXiv:1506.02142},
  2015.

\bibitem{oquab2014learning}
M.~Oquab, L.~Bottou, I.~Laptev, and J.~Sivic, ``Learning and transferring
  mid-level image representations using convolutional neural networks,'' in
  \emph{Proceedings of the IEEE conference on computer vision and pattern
  recognition}, 2014, pp. 1717--1724.

\bibitem{simonyan2014very}
K.~Simonyan and A.~Zisserman, ``Very deep convolutional networks for
  large-scale image recognition,'' \emph{CoRR}, vol. abs/1409.1556, 2014.

\bibitem{sharif2014cnn}
A.~Sharif~Razavian, H.~Azizpour, J.~Sullivan, and S.~Carlsson, ``Cnn features
  off-the-shelf: an astounding baseline for recognition,'' in \emph{Proceedings
  of the IEEE Conference on Computer Vision and Pattern Recognition Workshops},
  2014, pp. 806--813.

\bibitem{stack2011automation}
J.~Stack, ``Automation for underwater mine recognition: current trends and
  future strategy,'' in \emph{SPIE Defense, Security, and Sensing}.\hskip 1em
  plus 0.5em minus 0.4em\relax International Society for Optics and Photonics,
  2011, pp. 80\,170K--80\,170K.

\bibitem{zhu2014model}
Z.~Zhu, X.~Xu, L.~Yang, H.~Yan, S.~Peng, and J.~Xu, ``A model-based sonar image
  atr method based on sift features,'' in \emph{OCEANS 2014 - TAIPEI}, April
  2014, pp. 1--4.

\bibitem{wright2009robust}
J.~Wright, A.~Y. Yang, A.~Ganesh, S.~S. Sastry, and Y.~Ma, ``Robust face
  recognition via sparse representation,'' \emph{Pattern Analysis and Machine
  Intelligence, IEEE Transactions on}, vol.~31, no.~2, pp. 210--227, 2009.

\bibitem{mckay2015discriminative}
J.~McKay, R.~Raj, V.~Monga, and J.~Isaacs, ``Discriminative sparsity in sonar
  atr,'' \emph{Oceans 2015 Washington, DC}, 2015.

\bibitem{fandos2011optimal}
R.~Fandos and A.~M. Zoubir, ``Optimal feature set for automatic detection and
  classification of underwater objects in sas images,'' \emph{Selected Topics
  in Signal Processing, IEEE Journal of}, vol.~5, no.~3, pp. 454--468, 2011.

\bibitem{kumar2012object}
N.~Kumar, Q.~F. Tan, and S.~S. Narayanan, ``Object classification in sidescan
  sonar images with sparse representation techniques,'' in \emph{Acoustics,
  Speech and Signal Processing (ICASSP), 2012 IEEE International Conference
  on}.\hskip 1em plus 0.5em minus 0.4em\relax IEEE, 2012, pp. 1333--1336.

\bibitem{mckay2016localized}
J.~McKay, V.~Monga, and R.~Raj, ``Localized dictionary design for geometrically
  robust sonar atr,'' \emph{IGARSS 2016}, 2016.

\bibitem{mckay2016robust}
------, ``Robust sonar atr with pose corrected sparse reconstruction-based
  classification,'' \emph{OCEANS 2016 MTS/IEEE Monterey}, Sept 2016.

\bibitem{krizhevsky2012imagenet}
A.~Krizhevsky, I.~Sutskever, and G.~E. Hinton, ``Imagenet classification with
  deep convolutional neural networks,'' in \emph{Advances in neural information
  processing systems}, 2012, pp. 1097--1105.

\bibitem{wang2016cnn}
J.~Wang, Y.~Yang, J.~Mao, Z.~Huang, C.~Huang, and W.~Xu, ``Cnn-rnn: A unified
  framework for multi-label image classification,'' \emph{arXiv preprint
  arXiv:1604.04573}, 2016.

\bibitem{lee2011unsupervised}
H.~Lee, R.~Grosse, R.~Ranganath, and A.~Y. Ng, ``Unsupervised learning of
  hierarchical representations with convolutional deep belief networks,''
  \emph{Communications of the ACM}, vol.~54, no.~10, pp. 95--103, 2011.

\bibitem{chatfield12014return}
K.~Chatfield, K.~Simonyan, A.~Vedaldi, and A.~Zisserman, ``Return of the devil
  in the details: Delving deep into convolutional nets,'' in \emph{British
  Machine Vision Conference}, 2014.

\bibitem{vedaldi2015matconvnet}
A.~Vedaldi and K.~Lenc, ``Matconvnet: Convolutional neural networks for
  matlab,'' in \emph{Proceedings of the 23rd ACM international conference on
  Multimedia}.\hskip 1em plus 0.5em minus 0.4em\relax ACM, 2015, pp. 689--692.

\bibitem{zeiler2014visualizing}
M.~D. Zeiler and R.~Fergus, ``Visualizing and understanding convolutional
  networks,'' in \emph{European conference on computer vision}.\hskip 1em plus
  0.5em minus 0.4em\relax Springer, 2014, pp. 818--833.

\end{thebibliography}

%% End Document
\end{document}